\documentclass{article}
 
\usepackage{arxiv}
\usepackage[utf8]{inputenc}
\usepackage{subfig}
\usepackage{graphicx}
\usepackage{microtype}
\usepackage{booktabs} 
\usepackage{amsmath}
\usepackage{booktabs}
\usepackage{amssymb}
\usepackage[pagebackref=true,breaklinks=true,colorlinks,bookmarks=false]{hyperref} 

\begin{document}

\title{Saliency Map Based Data Augmentation}

\author{Jalal Al-afandi, Bálint Magyar, András Horváth\\
Peter Pazmany Catholic University Faculty of Information Technology and Bionics\\
Budapest, Práter u. 50/A, 1083\\
{\tt\small alafandi.mohammad.jalal, magyar.balint, horvath.andras@itk.ppke.hu}
}

\maketitle

\begin{abstract}
Data augmentation is a commonly applied technique with two seemingly related advantages. With this method one can increase the size of the training set generating new samples and also increase the invariance of the network against the applied transformations.
Unfortunately all images contain both relevant and irrelevant features for classification therefore this invariance has to be class specific. 
In this paper we will present a new method which uses saliency maps to restrict the invariance of neural networks to certain regions, providing higher test accuracy in classification tasks.
\end{abstract}

\section{Introduction}

The accuracy and practical performance of machine learning methods depend heavily on the collected datasets. 
In case of complex problems extremely large datasets are needed and not only their collection but also their annotation is a cumbersome and tedious job.
Various approaches have appeared in the past to overcome or ease this problem such as transfer learning \cite{torrey2010transfer}, few-shot learning \cite{snell2017prototypical}, training on synthetic data \cite{tremblay2018training} and domain adaptation \cite{ganin2016domain}, but the most commonly applied method is still data augmentation \cite{shorten2019survey}.

In case of data augmentation transformations are used in the dataset meanwhile the labels are also adjusted accordingly (typically kept the same in case of classification). This way one can generate new data-points which correlate with the original one, but might contain additional information. For example in case of a dog/cat classifier one can rotate, shift and scale the images and by this creating new samples which can be considered new dog/cat images which could have appeared during data collection. This augmentation strategy when new samples are generated with the same labels can be referred to as output class based data augmentation.

Another viewpoint of data augmentation is the increased feature invariance of the network.
One can imagine that a small shift, rotation, noise etc. should have a negligible effect on the output of the network and expect that the application of these transformations in the training set increases the invariance of the network.
Unfortunately this is not necessarily true, as it was demonstrated in \cite{azulay2018deep}, \cite{zhang2019making} and \cite{hernandez2019learning} that convolutional networks are sensitive towards these perturbations, even when they were used in training with class based data augmentation.
To further increase the invariance of these networks an auxiliary loss function can be applied \cite{hernandez2019learning}, which ensures that the original and the transformed input samples result the same output in the logit layer or even the same embedded representation at selected layers. This strategy could be referred to as invariance based or feature level data augmentation.
We have to mention that although invariance based data augmentation is commonly applied in practice and can significantly increase the accuracy and the invariance of the networks \cite{hernandez2019learning} it can be difficult to identify which layer is optimal to calculate the difference between the original and transformed responses to expect invariant representations.

These two approaches can contradict to each other in certain cases, meanwhile both of them are partially substantiated in the literature.
There might be certain transformations which can change the output confidence or class of a certain object. For example too much rotation could result an object which will never appear in practice (e.g it is a fair assumption that lamp posts are always upright and vertical on images and one could expect lower logit values to a rotated lamp post), meanwhile  it is also a fair assumption that a network, apart from the object class, should detect these small transformations and this goes against class based data augmentation.
There might be some features which are important for a specific class (e.g a tilted or upright head of a dog or person could signify different emotions in certain applications) and it might not be advantageous in some classes if the network encodes exactly the same activations for the original and transformed input samples.

\begin{figure}[htp]
 \centering
\subfloat{   \includegraphics[width=1.7in]{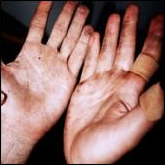}}
\subfloat{   \includegraphics[width=1.7in]{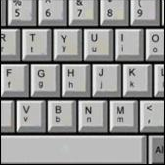}}
\caption{Two selected images from ImageNet representing the problem of feature invariance in general. The left image is labeled as band aid and the right image as space bar. The objects determining the output class in this task occupy only a small area of the input and this can cause problems in case of invariance based data augmentation. The second image also demonstrates how invariance could be task dependent. For example the orientation of a keyboard and the space bar could be arbitrary, but if the same image is used for letter recognition the orientation of the letters can be important and upside down letters are typically uninterpretable.}
\label{fig:InvarianceSample}
\end{figure}

In every input image there might be important regions which could modify the certainty of the output class and it can be important to reflect the transformation of these regions, meanwhile there might be background and unimportant regions which are irrelevant for the specific class and the network output should be irrelevant to these changes. Figure \ref{fig:InvarianceSample} illustrates this problem with two samples taken from ImageNet.

There are approaches where saliency maps are used to select regions in the image for data augmentation \cite{gong2021keepaugment} or where the augmentation strategies are tuned according to the saliency maps\cite{dabouei2021supermix}, but none of these approaches enforce invariance on the network using saliency maps, these methods only help combining multiple images.

In this paper we will present a novel method which can be applied with all known data augmentation techniques and instead of the application of auxiliary loss functions at specific layers which would force invariant features on the network, we encourage the network to focus on the same regions during classification for the original and augmented samples, by comparing and calculating a loss between their saliency maps.

\section{Saliency Maps}\label{Sec:Saliency_literature}
The goal of using saliency maps (also referred to as sensitivity map or pixel attribution map in literature) is to provide a visual explanation about which regions of the input image contribute to the network's decision.

The first method was introduced by Simonyan et al. \cite{simonyan2013deep}. It is based on simple \textbf{gradient backpropagation}. In contrast to the training where we calculate the gradients of the loss function with respect to the network parameters, here the gradients of the predicted class score ($S_c$) with respect to the input image pixels ($I$) are calculated:

\begin{equation}
   M(I) = \frac{\partial}{\partial I} S_c
\end{equation}

The dimensions of the resulting saliency map $M$ will be equal to the dimensions of the input image, for an RGB image  $M \in \mathbb{R}^{m\times n \times 3}$. In order to visualize $M$, it needs to be converted to a single channel map $\hat{M}$.

In the original paper, the authors took the highest value across each channel in every $ij$ pixel position: 

\begin{equation}
   \hat{M_{ij}} = \max_c(M_{ijc})
\end{equation}

\begin{figure}[htp]
 \centering
\subfloat{\includegraphics[width=3.5in]{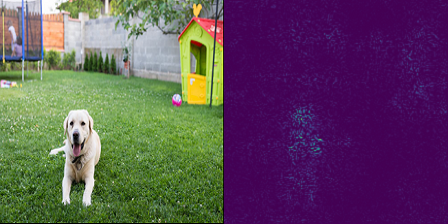}}
\caption{Saliency visualization using the gradient based method from \cite{simonyan2013deep}. The network predicted the right class, Labrador Retriever with 0.71 confidence. The input image is displayed on the left and the weights of the pixels in the network decision is depicted on the right. }
\label{fig:saliency_example}
\end{figure}

Figure \ref{fig:saliency_example} shows an example of this saliency visualization method.

There are two, very similar, methods to gradient backpropagation \textbf{DeconvNet} \cite{zeiler2014visualizing} and \textbf{guided backpropagation} \cite{springenberg2014striving}. Both methods start from the target class activation and calculate the saliency map by applying the reverse operation of all layers (convolution, pooling, activation), and by backpropagation.

The only significant difference between these three methods, is how the ReLU activation is handled during saliency map calculation. Feature map $f^{l+1}$ activation after ReLU is calculated as follows:

\begin{equation}
   f^{l+1} = \max_c(f^l,0)
\end{equation}

In the backward pass, the following functions are used to calculate the reconstructed layer $R^l$ for the backpropagation \cite{simonyan2013deep}, DeconvNet \cite{zeiler2014visualizing} and guided backpropagation \cite{springenberg2014striving} methods respectively:

\begin{equation}
\begin{aligned}
   R^l_{backpr} &= (f^l>0)*R^{l+1} \\
   R^l_{deconv} &= (R^{l+1}>0)*R^{l+1} \\
   R^l_{guided} &= (f^l>0)*(R^{l+1}>0)*R^{l+1} \\
\end{aligned}
\end{equation}

An improvement of the gradient based methods is the \textbf{SmoothGrad} technique \cite{smilkov2017smoothgrad}. The authors state that  the derivative of $S_c$ fluctuates greatly at small scales, gradients are not smooth and local variations of the derivatives introduce noise in the saliency map.

This noise can be eliminated by applying the following steps:
\begin{itemize}
\item Generating $n$ number of images from the original image by adding Gaussian noise.
\item Calculating the saliency map for each image using gradient backpropagation.
\item Averaging the calculated saliency maps.
\end{itemize}

\textbf{Grad-CAM} \cite{selvaraju2017grad} that stands for Gradient-weighted Class Activation Mapping is one of the most popular saliency techniques.

\begin{figure}[htp]
 \centering
\subfloat{\includegraphics[width=3.5in]{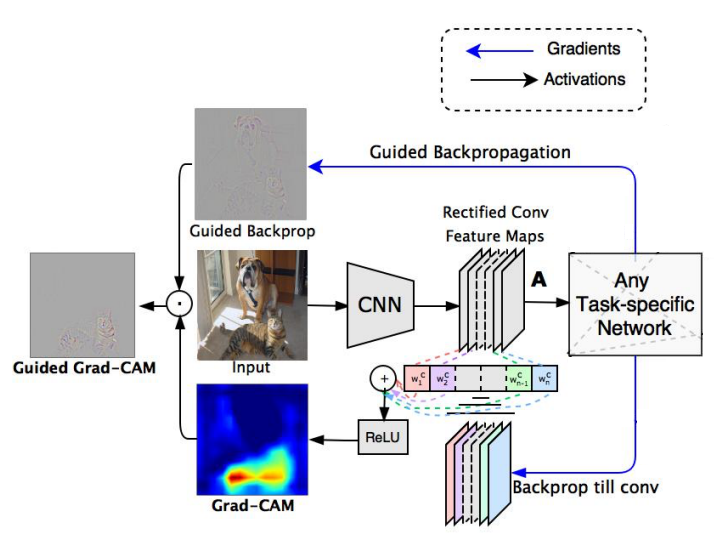}}
\caption{Saliency visualization using the Gradient-weighted Class Activation Mapping method from \cite{selvaraju2017grad}. }
\label{fig:gradcam_model}
\end{figure}

Grad-CAM has two pathways to calculate the gradients (its computation is depicted in Figure \ref{fig:gradcam_model}). The first one uses guided-backpropagation \cite{springenberg2014striving} to calculate a saliency map.
In the second path, the gradient of the target class activation is propagated back only until the last convolutional layer. Then, the values of this convolutional layer's feature maps are weighted by the backpropagated gradient. After that an average feature map is generated and only the positive values are kept by applying ReLU activation. This average feature map will be resized to the input image size, and merged with the saliency map from the first pathway.

\vspace*{10px}

In our method we wanted to use the simplest, computationally efficient saliency map technique, therefore we applied the backpropagation based method from \cite{simonyan2013deep}.

\section{Data Augmentation}\label{Sec:DataAug}

\subsection{Data Augmentation Methods}

Data augmentation is commonly used in machine learning and its effect was demonstrated on various datasets \cite{shorten2019survey}.
There are numerous methods which might depend on the specific task and also on the domain of the data.
Here we will list strategies and methods which are related to computer vision, but data augmentation is not at all limited to images. It is commonly put into effect in natural language processing, speech processing, time series analysis and other applications and the used methods could vary accordingly.

The first applied approaches were traditional image transformation methods where the shape and quality of the images were transformed. The transformation methods included rotations, shifts and scaling the image with affine or even elastic transformations or changing the colors of the image by modifying the histogram or switching color channels \cite{mikolajczyk2018data}.

Later more complex methods combining multiple images have appeared, such as MixUp \cite{zhang2017mixup}, CutOut \cite{devries2017improved} or AugMix \cite{hendrycks2019augmix}. In these approaches multiple images (in practice mostly two) are combined using opacity and/or cutting and pasting random regions. The labels are also combined in a weighted manner according to the opacities and the areas used from the two images.

A fairly comprehensive overview of recent data augmentation methods along with their advantages can be read in \cite{shorten2019survey} also a wide range of collection can be found under \href{https://github.com/AgaMiko/data-augmentation-review
}{this link}.

The comparison of these many methods is difficult in practice especially since, as we argued earlier the proper augmentation methods and their parametrization can depend on the image content as well. Because of this the optimal augmentation strategy was also investigated and optimized as a reinforcement learning task which created AutoAugment \cite{cubuk2019autoaugment}, where an algorithm tries to select the most suitable transformations for every input sample based on their features.

\subsection{Data Augmentation Strategies}
In standard data augmentation methods some transformations are applied on the input images and the label is kept unchanged or transformed accordingly. In these scenarios we hypothesize a transformation which could happen in the wild, during data acquisition and we generate new data points as inputs for the same class. In classification problems we expect the network to give the same output for the original and the transformed images:

\begin{equation}
   L_{class} = L_x (N_k(x), l_x ) + L_x ( N_l(T(x) , l_x)
\end{equation}
Where $N_k$ represents the  activations in the last, logit layer of our neural network (with layers from $1$ to $k$), $x$ is our input sample, $T$ is the transformation used for data augmentation, $l_x$ is the ground truth label of the sample and $L_x$ is the applied loss function.
In detection and segmentation problems the ground truth labels are also transformed in the same manner as the input: $T(l_x)$.

We see these strategies as a label oriented data augmentation, which focuses on the logit layer of the network, the inner representation is not important at all, the only important aspect is that the original and the transformed samples should result the same output class.

It was demonstrated that this method on its own does not necessarily causes invariance\cite{hernandez2019learning}.
It was later observed that these transformations might be different for different classes, for example pencils and mobile phones could usually be in arbitrary orientation on images, meanwhile cars are typically in an upright position.

In other strategies one expects invariance for the network which can also mean that the inner representations are forced to be the same.
For example after scale change or additive noise, not only the output class, but also the activations in the $i$th layer has to be the same. This forces invariance directly on the network but unfortunately it is difficult to identify where this invariance or equivariance is optimal. Invariance based data augmentation can be formulated as:
\begin{equation}
  L_{inv} = \sum_{i \in H_T}  \left \| N_i(x), N_i(T(x))\right \|
\end{equation}
Where $\left \|y,z \|\right \|$ is a distance metric (typically $\ell_1$ or $\ell_2$ distance between $y$ and $z$) and $H_T$ is the set of layers in which the invariance is enforced for transformation T.

Invariance based loss can be used in an unsupervised manner, since as it can be seen from the equation it does not require the class labels, it only enforces invariance, but does not solve the problem of classification. Because of this invariance based loss is typically applied as an auxiliary loss, an extra regularizer along with the class based loss.

It is also a problem that invariance based loss is computed for all activations and for the whole image. For a specific input image it might be important to detect the change of certain regions, meanwhile simultaneously the same changes should be completely neglected for other regions of the same transformation.


\section{Saliency Map Based Data Augmentation}\label{Sec:Saliency}

As we have demonstrated it is difficult to identify which factors are important in data augmentation.
Should a new sample  be completely invariant for the transformation and result the same activations or should it be equivariant, representing the same class but also encoding the transformation.
This is an extremely difficult task and an image dependent question, which can not really be answered using the previously introduced strategies.
Here we would suggest a third approach instead of using data augmentation with the same labels or applying data augmentation directly on midway activations to enforce feature invariance.
We apply saliency map based data augmentation in which we introduce a third level of invariance. Apart form the activation level invariance and the class invariance of the network we introduce saliency map level invariance.
Saliency maps are traditionally generated for classification problems to visualize the important regions and factors for the decision of the network as it was described in section \ref{Sec:Saliency_literature}.
These approaches will highlight by a weighted map which pixels and regions were involved in the decision of the network.
We hypothesize that saliency maps are directly connected to network invariance in data augmentation.
If we present a transformed sample to the network one can not expect to result exactly the same output class (the augmentation might be too severe, for example an upside-down car might be suspicious) also one can not either expect to generate exactly the same activations, since it might be important for the network, but one could expect that in case of transformed samples (which contain the same objects) the network should focus on the same regions, the decision of the network should come from the same pixels, meanwhile the inner activations and representations could vary.

\begin{figure}[htp]
 \centering
\subfloat{   \includegraphics[width=3.0in]{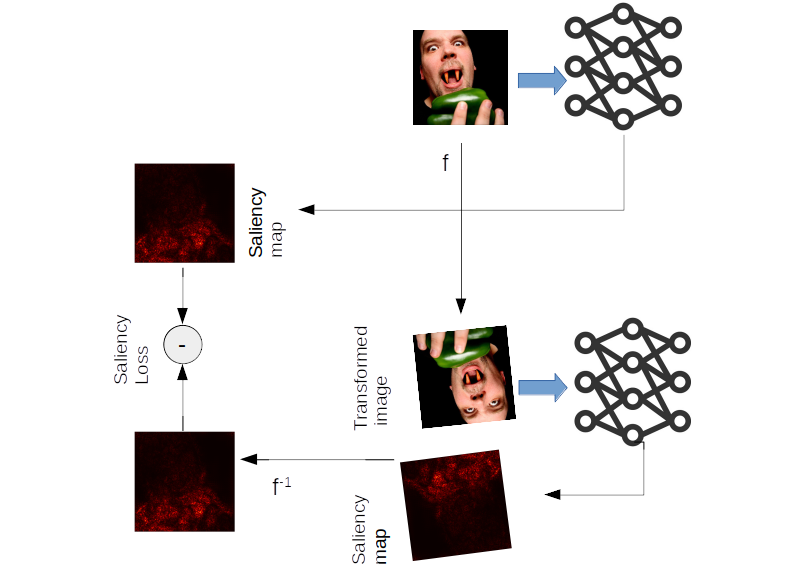}}
\caption{This figure depicts the calculation of saliency loss. We process original (undistorted) and augmented (transformed) images in the same batch, the parameters of the transformations are known. Once the inputs were processed by our neural network we calculate the saliency map for every image according to its output label. In the next step we apply the inverse of the transformation function on every augmented saliency map and compare them to their original counterpart.}
\label{fig:SaliencyLoss}
\end{figure}

Our method can be described in the following way:
we create a batch containing both original and transformed samples simultaneously. 
A forward and backward pass is implemented just as normally to calculate the gradients of the parameters for classification.
After the traditional backpropagation of the class labels, an extra backward pass is implemented to calculate the saliency maps for all input samples, each sample according to its desired output class.
The saliency maps of the original samples are selected and the maps of the transformed samples are transformed back, using the inverse transform of the augmentation function.
After this step we calculate a distance between the saliency maps of the original and transformed samples:

\begin{equation}
  L_{sal} = \left \| S(N_l(x)), T^{-1} ( S( N_l(T(x))))\right \|
\end{equation}
where $S$ calculates the saliency map of input $x$ for network $N_l$.
The distance metric between the two saliency maps could be an arbitrary metric such as $\ell_1$ or $\ell_2$. In our experiments we have normalized every saliency map individually and calculated their mean squared distance after normalization. A simple diagram depicting the calculation of the Saliency map based loss can be seen in Figure \ref{fig:SaliencyLoss}.

With this method instead of forcing the network for invariant representation at all 
or directly forcing the network for the same activations and by this enhancing invariance, 
we force the network to generate the same saliency maps and focus on the same features.

\begin{figure}[htp] 
\centering
\includegraphics[width=3.2in,height=1.7in]{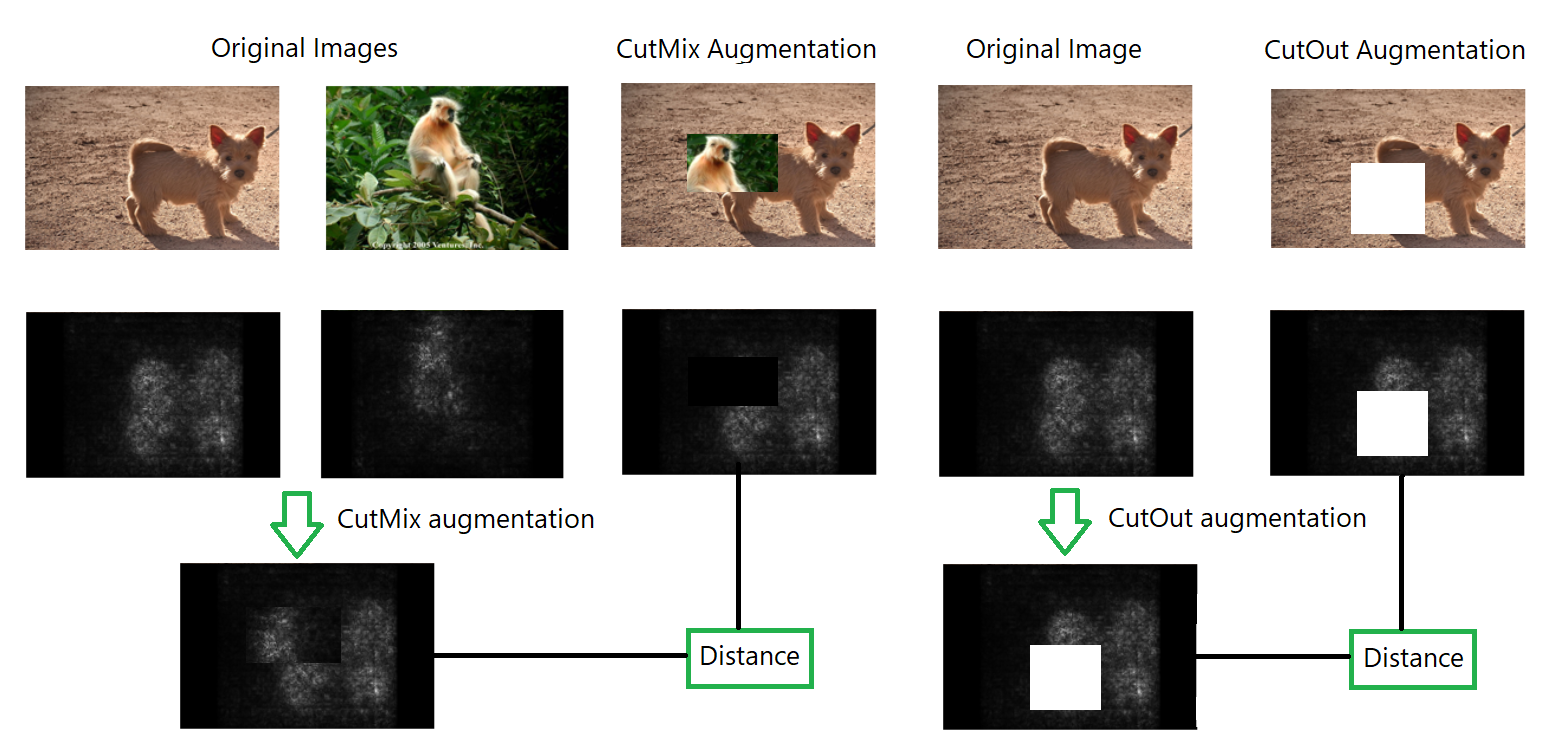}
\caption{It is an illustration for the feasibility of using saliency loss with CutOut and CutMix data augmentation. A similar approach can be adopted for other complex data augmentation methods where we compare the saliency map of the augmented image with the augmented version of the saliency map of the original image.}
\label{fig:augcutmix}
\end{figure}

Here we have described data augmentation as a method which transforms a singe sample ($T(x)$). As it was mentioned earlier there are more complex methods which combine multiple images, such as CutMix or MixUp. Since these methods apply simple transformations, the same functions can also be applied on the individual saliency maps to generate an expected map. For example in case of CutMix the parts of the original saliency maps can be cut and mixed and in case of MixUp they can be blended together with the same opacity.
An image depicting the steps of saliency map based loss for CutOut and CutMix augmentations can be seen in Figure \ref{fig:augcutmix}.

Similarly as invariance based loss can be used together with the class based loss function we also think about our method as an auxiliary loss function.
The three losses can be used together or separately and they can be arbitrarily weighted with the $\alpha$, $\beta$ and $\gamma$ parameters:
\begin{equation}
  L=  \alpha L_{class} + \beta L_{sal} + \gamma L_{inv} 
\end{equation}


\section{Experiments and Results}\label{sec:experiments}

Since there is a wide spectrum of data augmentation methods used in practice, we have selected  simple datasets such as CIFAR-10\cite{Cifar10} and SVHN\cite{sermanet2012convolutional} to demonstrate the effectiveness of our approach.
The images of the datasets were rescaled to 224x224 and investigated with various network architectures. We have used the method described in \cite{simonyan2013deep} to calculate the saliency maps. 

In our experiments we identified two different dimensionalities for comparison. The first was the comparison of different data augmentation methods (e.g scaling, rotation, Mixup etc) and the second was the strategy applied for data augmentation (class, invariance and saliency based augmentation). Of course these two factors are not independent from each other but combining both would result an extremely high number of parameters.

In the following we will list the main parameters of our experiments. For the detailed set of parameters and implementation of our experiments  please check our code in the supplementary material.

\subsection{Comparing Augmentation Methods}

First we compared the effect of saliency based data augmentation with various augmentation methods.

We used the class based loss as a baseline and for comparison we have included the saliency loss during training as a regularizer to force the network to concentrate on the same pixels of the original and transformed samples. 

We have trained a ResNet-18 architecture on the scaled images of the CIFAR-10  and SVHN datasets and compared the test accuracies of the network. The average test accuracies of five independent trainings using various augmentation methods can be found in Table \ref{tab:AugmentaitonMethods}.

There was a fifty percent chance that an image will be vertically mirrored and also the same chance that its contrast will be enhanced, similarly the image rotated, scaled, horizontally mirrored and the contrast was changed randomly with the parameters described in the caption of Table \ref{tab:AugmentaitonMethods}.

\begin{table}[h!]
\centering
\begin{tabular}{@{}l c c c@{}}

\textbf{ResNet-18} &  SVHN & CIFAR10 \\ \midrule
\hline
HorFlip Ref&  $91.5\%$ & $71.9\%$    \\
HorFlip Sal&  $91.5\%$  &   $73.4\%$    \\
\hline
VertFlip Ref&  $89.4\%$  &    $76.3\%$    \\
VertFlip Sal &  $92.1\%$  &   $75.3\%$    \\
\hline
Rescale Ref&  $92.1\%$  &  $74.2 \%$   \\
Rescale Sal&  $92.3\%$  &  $73.6 \%$   \\
\hline
Contrast Ref& $92.3\%$   &   $73.5 \%$    \\
Contrast Sal&  $92.1\%$  &    $73.1\%$    \\
\hline
Rotation Ref&   $91.6 \%$ &  $74.5\%$  \\
Rotation Sal &  $91.9 \%$ &   $75.4\%$  \\
\hline
Combined Ref&  $92.2\%$  &    $73.7\%$   \\
Combined Sal &  $92.8\%$  &   $74.0  \%$   \\
\hline
MixUp Ref\cite{zhang2017mixup} &   $91.7\%$  &   $74.8\%$  \\
MixUp Sal & $92.1\%$ &   $75.4\%$  \\
\hline
CutOut Ref\cite{devries2017improved} o & $91.5\%$ &  $74.9\%$ \\
CutOut Sal & $92.1\%$ &   $76.2\%$    \\
\hline
AugMix Ref\cite{hendrycks2019augmix} &  $92.1\%$   &  $74.8\%$    \\
AugMix Sal & $92.5\%$  &   $75.4 \%$ \\
\hline
\end{tabular}
\caption{This table depicts the test accuracies of five independent network trainings on the SVHN and CIFAR-10 datasets using the ResNet-18 architecture with traditional and saliency based  data augmentation. Each pair of rows list a selected data augmentation method for the traditional (Ref using class based data augmentation) and saliency based data augmentation methods (Sal). The investigated augmentation methods were the following:
HorFlip - mirroring the images horizontally, VertFlip - mirroring the images vertically, Rescale- randomly magnifying the image with a maximum of ten percent, Contrast - Contrast Limited Adaptive Histogram Equalization,  Rotation- random rotation with a maximum of ten degrees, Combined - a combination of all previously listed methods.}
\label{tab:AugmentaitonMethods}
\end{table}

As it can be seen from the results saliency based data augmentation has increased the test accuracy in almost all cases for both datasets. The increase in accuracy is not drastic but consistent and could be important in practical applications.

One can also see that the best performing data augmentation methods are the combination of multiple simple transformations (Affine transformation) or novel and complex augmentation methods such as AugMix, MixUp or CutOut.

We also have to note that saliency based loss was used during the whole training of the networks. It is easy to see that saliency maps are useful once the network has learned distinguishing features, but we have applied the saliency map based loss even in the first iterations, when the convolutional kernels were initialized randomly.

\subsection{Comparing Augmentation Strategies}

Based on the results of the previous subsection summarized in Table \ref{tab:AugmentaitonMethods} we have selected an affine transformation which we adopted from the semi-supervised paper \cite{chen2020simple} as a complex augmentation method and used it for detailed comparison.

In our implementation the transformations were executed randomly and independently from each other. In these comprehensive experiments we used crop and rescale with color jitter transformation as described in \cite{chen2020simple}. 

We have selected more complex architectures and investigated the effect of this transformation on SVHN, CIFAR-10 and also a more complex dataset, CIFAR-100.

The test accuracies for the average of three independent runs can be seen in Table \ref{tab:ArchitectureComp}. 

\begin{table}[h!]
\centering
\begin{tabular}{@{}lccc@{}}
\hline
   \textbf{Architecture} &  SVHN & CIFAR10 & CIFAR100\\ \midrule
\hline
DenseNet-121 Ref &  $76.4\%$   &    $78.7\%$ &   $52.6\%$   \\
DenseNet-121 Sal &  $80.8\%$ &     $79.1\%$  &   $52.9\%$     \\
\hline
ResNet-18 Ref&  $89.1\%$ &   $78.8\%$    &   $51.5\%$    \\
ResNet-18 Sal &  $89.5\%$ &   $79.2\%$    &  $52.7\%$       \\
\hline
\end{tabular}
\caption{The table contains the test accuracies where different rows belong to different network architectures and the columns contain three different datasets. Both architectures were trained with the traditional data augmentation methods (Ref) and our saliency map based approach (Sal)}
\label{tab:ArchitectureComp}
\end{table}

As it can be seen from the results saliency based data augmentation yields higher training accuracies in all investigated cases.



\subsection{Results on ImageNet}

As a larger scale study we investigated the effect of saliency map based data augmentation on the ImageNet Large Scale Visual Recognition Challenge 2012 Dataset. We have applied affine transformation as a data augmentation method (with the parameters described in Table \ref{tab:AugmentaitonMethods}). We have investigated the VGG-16 and 19 architectures with batch normalization\cite{simonyan2014very}, ResNet-50\cite{he2016deep} and DenseNet-121 and 169\cite{huang2017densely}.
We have compared the accuracies without data augmentation, with the combination of the class based and invariance based losses (with 0.5 and 0.5 weights) and a version where all loss functions, class based, invariance based and saliency map based losses were utilized with uniform weights.
The top-1 test accuracy results can be seen in Table \ref{tab:IamgeNetAcc}.

\begin{table}[h!]
\centering
\begin{tabular}{@{}l c c c @{}}
\hline
   \textbf{Architecture} &  No Aug & Class+Inv & Class+Inv+Sal\\ \midrule
\hline
VGG-16& $70.8\%$  &  $74.4\%$   &   $76.1\%$   \\
VGG-19 & $71.3\%$ &  $74.5\%$    &   $76.3\%$    \\
ResNet-s50& $72.1\%$  &   $78.4\%$  &   $79.2\%$\\
DenseNet-121 &  $75.1\%$ & $78.7\%$    &   $79.8\%$      \\
DenseNet-169 & $76.4\%$  &  $79.1\%$   &   $80.4\%$      \\
\hline
\end{tabular}
\caption{This table contains the test accuracies on the test set of the ImageNet dataset. Different rows belong to different architectures, meanwhile the columns contain three different augmentation strategies. No Aug contains the accuracies without any augmentation at all, in the second column results utilizing the class and invariance losses can be found, meanwhile the last column contains when all three different loss functions were utilized together with uniform weights.}
\label{tab:IamgeNetAcc}
\end{table}

As it can be seen from the results saliency map based data augmentation resulted higher test accuracies for all tested architectures. A consistent increase can be observed compared to the best traditional data augmentation we could reach. The average increase of the test accuracy was $1.3\%$ with a variance of 0.17.

\subsection{Running times}

As it was demonstrated by the previous results saliency based data augmentation has viability in network training. It can increase the test accuracy of the networks, since it helps focusing on the important objects, regions and features on every images apart from the transformations used during data augmentation. Our method can be used with arbitrary network architectures, training methods and additional regularizers such as dropout or batch-normalization. 

Unfortunately the application of this method adds a significant extra computation during training.
The calculation of the saliency maps for every ground truth label is a complex computation and requires additional time during training.
Fortunately this does not prevent its application in practice, since training has to be done only once.
For a fair comparison we have measured the training time using traditional and saliency map based data augmentation for some commonly applied network architectures and the training time for every iteration can be seen in Table \ref{tab:Time}. The training time of the second 100 batches were measured and averaged to ignore the inaccuracies caused by GPU warm-up. In case of the traditional approach the calculation of the forward pass and the gradient computation was measured and in case of the saliency map based augmentation the forward pass, the saliency map calculation, the transformation of the saliency map, the difference of the saleincy maps and the gradient calculation were included. We did not consider the data loading and the image transformation times during training, since these are the same for both methods and are usually implemented using separate threads.
The measurements were done in Pytorch using an NVIDIA RTX-2080 TI GPU.

\begin{table}[h!]
\centering
\begin{tabular}{@{}lcc@{}}
\hline
   \textbf{Architecture} &  Baseline time  & Saliency time\\ \midrule
\hline
VGG-16& 152.23 ms  &  1176.71 ms \\
VGG-19 & 163.73 ms  &  1194.62 ms   \\
ResNet-50&  307.57 ms  & 2156.14 ms  \\
Densenet-121 &  312.52 ms &  2685.33 ms  \\
Densenet-169 &  406.54 ms  &  3789.62 ms    \\
\hline
\end{tabular}
\caption{Training times for difference architectures and for the traditional class based augmentation and our approach.}
\label{tab:Time}
\end{table}

As it can be seen from the results saliency based data augmentation requires a significant amount of additional computation, but network training is a computationally extensive process in itself and the inference time of a neural network is the most determining factor in a practical application, which remained the same.

\section{Conclusion}
In this paper we have presented a new generic strategy for data augmentation, which utilizes the saliency maps during training to identify the important regions for invariance calculation and can be used with arbitrary network architectures or data augmentation methods. Our approach forces the network to focus on the same pixels and features on the original and transformed saliency images.
The only disadvantage of the method is that it requires extra computation during training, but it has no additional computation on network inference.
We have demonstrated that our method results an average increase of $1.3\%$ on ImageNet top-1 classification accuracy.
\section*{Acknowledgement} 

This research has been supported by the Hungarian Government by the
following grant: TKP2021\_02-NVA-27 – Thematic Excellence Program.

\bibliographystyle{ieeetr}
\bibliography{saliency.bib}

\end{document}